\documentclass[letterpaper, 10 pt, conference]{ieeeconf}
\IEEEoverridecommandlockouts
\usepackage{cite}
\usepackage{balance} % Balance columns at the end
\usepackage{moreverb,url}
\usepackage[colorlinks,bookmarksopen,bookmarksnumbered,citecolor=red,urlcolor=red]{hyperref}
\usepackage{siunitx}
\usepackage{xspace}
\usepackage{tabularx}
\usepackage{xcolor,colortbl}
\usepackage[table]{xcolor}
\usepackage{xparse}
\usepackage{mathtools}
\usepackage{url}
\usepackage{multicol}
\usepackage{soul}
\usepackage{color}
\usepackage{times}
\usepackage{graphicx}
\usepackage{lipsum}
\usepackage{setspace}
\usepackage{epsfig}
\usepackage{amsmath}
\usepackage{bm}
\usepackage{amssymb}
\usepackage{tabularx}
\usepackage{mathtools}
\usepackage{paralist}
\usepackage{tikz}
\usepackage{float}
\usepackage{pdfpages}
\usepackage{makecell}
\usepackage{array}
\usepackage{subcaption}
\usepackage{enumerate}
\usepackage{relsize}
\usepackage{pgfgantt}
\usepackage{amsfonts}
\usepackage{textcomp}
\usepackage{multirow,diagbox}
\usepackage{threeparttable}
\usepackage{booktabs}
\usepackage{bm}
\usepackage{booktabs}
\usepackage{tabularx}
\usepackage{amsmath}
\usepackage{siunitx}

\usepackage{{makecell}}
\usepackage{array,collcell}
\usepackage{algorithmic}
\usepackage{graphicx}
\usepackage{textcomp}
\usepackage{xcolor}
\def\BibTeX{{\rm B\kern-.05em{\sc i\kern-.025em b}\kern-.08em
    T\kern-.1667em\lower.7ex\hbox{E}\kern-.125emX}}

\newcommand{\etal}{\textit{et al}.~}
\def\figref#1{Fig.~\ref{#1}}
\def\tabref#1{Table~\ref{#1}}
\def\secref#1{Section~\ref{#1}}
\def\eqref#1{Eq.~(\ref{#1})}

\usepackage{pifont}
\newcommand{\cmark}{\ding{51}}
\newcommand{\xmark}{\ding{55}}

\begin{document}

% Types
\newcommand\todos[1]{\textcolor{red}{TODO: #1}} % TODO

%Commands for comments
\newcommand{\comments}[1]{\textcolor{red}{#1}}
\newcommand\Viorela[1]{\textcolor{red}{Viorela: #1}}
\newcommand{\Mik}[1]{\textcolor{magenta}{Mik: #1}}
\newcommand{\Jesse}[1]{\textcolor{orange}{Jesse: #1}}
\newcommand{\Yiduo}[1]{\textcolor{blue}{Yiduo: #1}}
\newcommand{\Desc}[1]{\textcolor{orange}{#1}}
\newcommand{\Summary}[1]{\textcolor{green}{#1}}

% Highlighting
\newcommand{\ToRemove}[1]{\textcolor{brown}{Remove: #1}}
\newcommand{\ToChange}[1]{\textcolor{olive}{Change: #1}}
\newcommand{\ToAdd}[1]{\textcolor{orange}{Add: #1}}
\newcommand{\ChangedTo}[2]{%
  \textcolor{purple}{Changed: }%
  \textcolor{olive}{"#1"}%
  \textcolor{red}{\ to:\ }%
  \textcolor{gray}{"#2"}%
}
\newcommand{\New}[1]{\textcolor{cyan}{New: #1}}

\newcommand{\Verify}[1]{\textcolor{blue}{#1}}

%Commands for special colours
\definecolor{niceorange}{RGB}{230,159, 0}
\definecolor{niceskyblue}{RGB}{86,180, 233}
\definecolor{nicebluishgreen}{RGB}{0,158, 115}
\definecolor{niceyellow}{RGB}{240,228, 66}
\definecolor{niceblue}{RGB}{0,114, 178}
\definecolor{nicevermillion}{RGB}{213,94, 0}
\definecolor{nicereddishpurple}{RGB}{204,121, 167}

\definecolor{light_blue}{HTML}{C5CAFB} 
\definecolor{light_red}{RGB}{255, 100, 100}

%%% Mik Colors
\definecolor{PriorFactor}{RGB}{0,0,0}
\definecolor{HybridMotionFactor}{RGB}{204,121,167}
\definecolor{ObjectSmoothingFactor}{RGB}{0,0,255}
\definecolor{OdometryFactor}{RGB}{255,153,51}
\definecolor{PointMeasurementFactor}{RGB}{150,150,150}
\definecolor{MotionModelFactor}{RGB}{255,221,85}
\definecolor{LimitFactor}{RGB}{85,221,255}
\definecolor{CostFactor}{RGB}{160,137,44}
\definecolor{ConstantAccFactor}{RGB}{128,255,128}
\definecolor{GoalFactor}{RGB}{120,68,33}
\definecolor{FollowFactor}{RGB}{212,0,0}
\definecolor{DynamicObstacleFactor}{RGB}{212,0,0}
\definecolor{StaticObstacleFactor}{RGB}{0,128,128}
\definecolor{LocalizationFactor}{RGB}{255,0,255}

\definecolor{DynamicObject}{RGB}{255,0,0}
\definecolor{EgoRobot}{RGB}{0,0,0}
\definecolor{EgoEstimation}{RGB}{0,158,115}
\definecolor{EgoPlan}{RGB}{211,141,0}
\definecolor{GlobalPlan}{RGB}{255,0,255} % Alpha 0.5
\definecolor{LocalGoal}{RGB}{212,102,217}
\definecolor{ObjectEstimation}{RGB}{123, 154, 239}
\definecolor{ObjectPrediction}{RGB}{0,38,147}

% From TRO for estimation
\newcommand{\inv}{^{-1}}
\newcommand{\tr}{^{\!\top}}
\newcommand{\invtr}{^{-\!\top}}
\newcommand{\invtrs}{^{-\!\top\slash2}}
\newcommand{\invs}{^{-1\slash2}}
\newcommand{\argmax}{\operatornamewithlimits{argmax}}
\newcommand{\argmin}{\operatornamewithlimits{argmin}}
\newcommand{\diag}{\mathit{diag}}
\newcommand{\se}{\mathrm{se}(3)}
\newcommand{\SE}{\mathrm{SE}(3)}
\newcommand{\SO}{SO(3)}
\newcommand{\R} {{\rm I\!R}}
\newcommand{\E} {{\rm I\!E}}
\newcommand{\bl}{{\bar l}}
\newcommand{\eqdef}{\vcentcolon=}
\newcommand{\algrule}[1][.5pt]{\par\vskip.5\baselineskip\hrule height #1\par\vskip.5\baselineskip}

\newcommand{\rot}[4]{\prescript{#1}{#3}{\mathbf{R}}^{#2}_{#4}}
\newcommand{\tran}[4]{\prescript{#1}{#3}{\mathbf{t}}^{#2}_{#4}}

\newcommand{\campose}[2]{\prescript{#1}{}{\mathbf{X}}_{#2}}
\newcommand{\objpose}[2]{\prescript{#1}{}{\mathbf{L}}_{#2}}
\newcommand{\worldf}{W}
\newcommand{\cammotion}[3]{\prescript{#1}{#2}{\mathbf{T}}_{#3}}
\newcommand{\objmotion}[3]{\prescript{#1}{#2}{\mathbf{H}}_{#3}}
\newcommand{\othmotion}[4]{\prescript{#1}{#2}{\mathbf{#3}}_{#4}}
\newcommand{\othpose}[3]{\prescript{#1}{}{\mathbf{#2}}_{#3}}
\newcommand{\objf}{L}
\newcommand{\camf}{X}
\newcommand{\imgf}{I}
\newcommand{\mpoint}[2]{\prescript{#1}{}{\mathbf{m}}_{#2}}
\newcommand{\nhpoint}[2]{\prescript{#1}{}{\tilde{\mathbf{m}}}_{#2}}
\newcommand{\ppoint}[2]{\prescript{#1}{}{\mathbf{p}}_{#2}}
\newcommand{\ipoint}[2]{\prescript{#1}{}{\mathbf{p}}_{#2}}
\newcommand{\icorre}[2]{\prescript{#1}{}{\tilde{\mathbf{p}}}_{#2}}
\newcommand{\opflow}[1]{\prescript{#1}{}{\bm{\phi}}}
\newcommand{\vel}[2]{\prescript{#1}{}{\mathbf{v}}_{#2}}
\newcommand{\acc}[2]{\prescript{#1}{}{\mathbf{a}}_{#2}}

\newcommand{\suchthat}{\;\ifnum\currentgrouptype=16 \middle\fi|\;}

\newcommand{\factor}[2]{\lVert {#1} 
\rVert^2_{\Sigma_{{#2}}}}
\newcommand{\ztwod}{\mathbf{z}_{\text{2D}}}
\newcommand{\zthreed}{\mathbf{z}_{\text{3D}}}

\title{\LARGE \bf DynoJEPP: Joint Estimation, Prediction and Planning\\in Dynamic Environments}

% \author{Mikolaj Kliniewski\mbox{*}\thanks{\mbox{*} Corresponding author}, Jesse Morris, Yiduo Wang, Ian Manchester and Viorela Ila% <-this % stops a space
% \thanks{The authors are with the Australian Centre for Robotics and School of Aerospace, Mechanical and Mechatronic Engineering, University of Sydney, Australia.
% 	{\tt \{mikolaj.kliniewski, jesse.morris, yiduo.wang, ian.manchester, viorela.ila\}@sydney.edu.au}}}

\author{Mikolaj Kliniewski\mbox{*}\thanks{\mbox{*} Corresponding author}, Jesse~Morris, Yiduo~Wang, Ian~R.~Manchester and~Viorela~Ila% <-this % stops a space
\thanks{Mikolaj Kliniewski, Jesse Morris, Yiduo Wang, Ian R. Manchester and Viorela Ila are with the Australian Centre For Robotics (ACFR), School of Aerospace, Mechanical and Mechatronic Engineering (AMME), University of Sydney, 2006 Sydney, Australia.
	{\tt \{mikolaj.kliniewski, jesse.morris, yiduo.wang, ian.manchester, viorela.ila\}@sydney.edu.au}}%
%\thanks{[\textbf{co}]: The two authors contributed equally to this work.}%
%\thanks{$^{\bf{*}}$\url{https://github.com/halajun/vdo_slam}}
}

\maketitle

% \begin{abstract} 
% Abstract
% \end{abstract}

% \begin{IEEEkeywords}
% keywords, keywords, keywords
% \end{IEEEkeywords}

% \section*{Conference Information}

% \noindent
% \textbf{ICRA 2026}\\
% \textbf{Deadline:}  September 15 - 23:59 PST.\\
% \textbf{Pages:} 8 (including all content: text, figures, tables, acknowledgments,  and references.)\\
% \textbf{Format:} double-anonymous (both reviewers and authors stay anonymous).\\
% \textbf{Accompanying video deadline:} September 17-22.

\begin{abstract} 
DynoJEPP is a factor-graph–based framework that jointly formulates and simultaneously optimizes estimation, prediction, and planning in dynamic environments. In conventional factor-graph-based approaches that jointly formulate estimation, prediction, and planning, information from prediction and planning feeds back into state estimation, yielding corrupted estimates, undesired behaviors, and unsafe plans.
%In conventional, factor-graph-based, jointly formulated approaches, information flows from prediction and planning back into estimation, leading to incorrect state estimates, undesired behaviors, and unsafe plans.
% We analyze prior joint optimization approaches and highlight the undesired information flow from future-related modules back to estimation. 
%To address this DynoJEPP introduces a novel \textit{directed factor} that enables directional influence within the factor graph.
%We evaluate its effect on the interaction of framework modules during navigation in static and dynamic environments.
%With directed factors, DynoJEPP traverses the environment safely; without them, it collides in most experiments.
%We further introduce Cooperative DynoJEPP, enabling the ego robot to account for cooperative object behavior in its prediction and trajectory planning.
To address this, DynoJEPP introduces a novel \textit{directed factor} that enforces directional information flow within the factor graph, preventing prediction and planning from corrupting state estimation. 
We evaluate the impact of directed factors on inter-module interactions during navigation in both static and dynamic environments. Our results demonstrate that these factors are critical for safe operation, as without them, the robot collides in the majority of experiments. Building on this, we further introduce Cooperative DynoJEPP, which enables the ego robot to incorporate cooperative object behavior into its prediction and trajectory planning.
% DynoJEPP will be open-sourced upon acceptance.}
% DynoJEPP is open-sourced at: link. 
% \comments{Remove open-sourced.}

% Problem and motivation - 1
% Key method 2
% Main results 2
% Contribution/Impact 1

\end{abstract}

% \vspace{1em}
% \textbf{Keywords:}

\section{Introduction}
% \textbf{1. Why we care about this topic? - broad motivation (real world) and specific problem. 2 + 3}

% Real world complication, need for motion understanding and prediction for planning

%The real world is rich with dynamic objects that exhibit complex motions, and poses challenges to robotic systems that need to navigate through such environments. 
Real-world robotic systems must operate in the presence of dynamic objects, where complex and unpredictable motions pose a fundamental challenge to safe and reliable autonomy.
% Understanding and predicting the motions of moving objects has been shown to improve robotic navigation and planning~\cite{Berg2006icra, Phillips2011ira_sipp, Hermann2014, Finean2023ras}, 
% producing safer and more optimal routes in dynamic scenes. 
% Many recent approaches consider both estimating robot poses based on historical observations and generating a safe future plan as variants of trajectory optimization problems~\cite{mukadam2017rss_STEAP, mukadam2019ar_STEAP}.
% Solving these problems as a unified system using factor graph reduces potential redundancies in the pipeline~\cite{mukadam2019ar_STEAP}, and taking into account estimations of both the static scene and dynamic object motions is highly relevant to planning optimal control actions~\cite{King2022ICRA_SCATE}.

%Many recent approaches for static environments treat both robot state estimation and path planning as variants of trajectory optimization problems~\cite{mukadam2017rss_STEAP, mukadam2019ar_STEAP}. 
%These can be solved in a unified framework using factor graphs, which reduces potential redundancies in the pipeline~\cite{mukadam2019ar_STEAP}.
%In dynamic environments, understanding and predicting the motions of moving objects is additionally important to further improve robotic navigation~\cite{Berg2006icra, Phillips2011icra_sipp, Hermann2014, Finean2023ras}. 
%Taking into account estimations of both the static scene and dynamic object motions is therefore highly relevant to planning optimal control actions~\cite{King2022ICRA_SCATE}.

In static environments, recent approaches formulate both robot state estimation and path planning as instances of trajectory optimization~\cite{mukadam2017rss_STEAP, mukadam2019ar_STEAP}.
These problems can be solved within a unified factor-graph framework, reducing redundancy across traditionally separated modules~\cite{mukadam2019ar_STEAP}.
However, real-world environments are inherently dynamic, where understanding and predicting the motion of surrounding objects, together with the static scene structure, becomes essential for reliable navigation~\cite{Berg2006icra, Phillips2011icra_sipp, Hermann2014, Finean2023ras, King2022ICRA_SCATE}.
While many solutions jointly formulate estimation and planning in static environments, their extension to dynamic scenes remains limited. As shown in~\tabref{tab:related_joint_work}, existing approaches typically either assume future object trajectories are known \textit{a priori}~\cite{King2022ICRA_SCATE} or rely on purely reactive avoidance strategies~\cite{Agha2018tro_SLAP}, restricting their applicability in realistic dynamic environments.
Moreover, simultaneously solving estimation and planning can cause the planning component to adversely influence state estimates themselves, leading to overly optimistic, and ultimately unsafe plans.

\begin{figure}[t]
    \centering
  \includegraphics[width=\columnwidth]{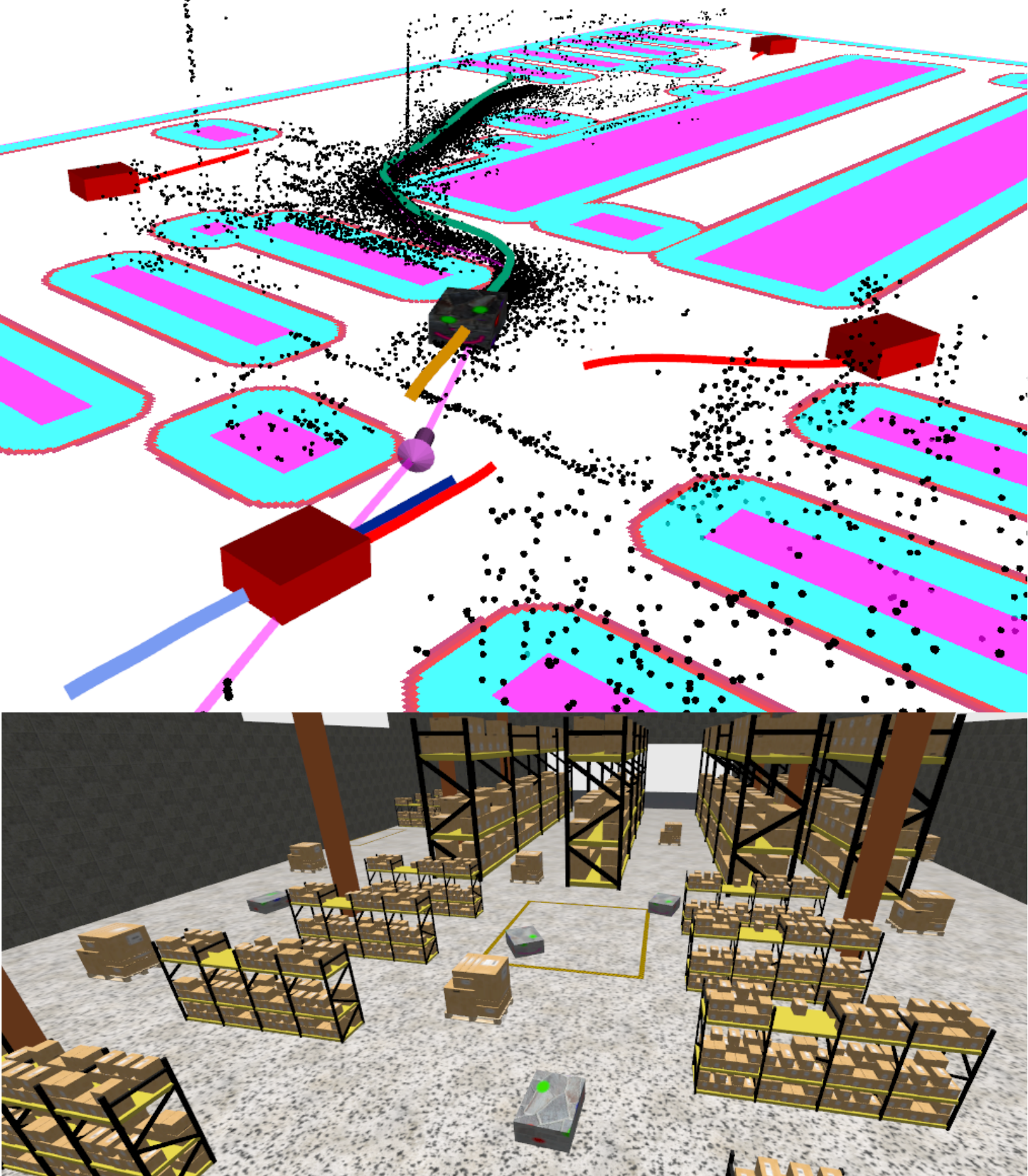}
   % \caption{\small{A robot navigating a simulated warehouse scenario with dynamic obstacles, i.e. other robots. The controlled robot is colored black while all other robots are red. A Signed Distance Field (SDF) map is generated from the \Verify{precomputed occupancy grid of the} static scene, and a global path is planned based on it. We \Verify{simultaneously} estimate the robot's pose and dynamic object motions, predict the trajectory of moving obstacles, and plan a safe local path to avoid both static and dynamic obstacles while being guided by a local goal on the global plan.}}
   \caption{\small{A robot navigating a simulated warehouse environment with dynamic obstacles. The ego robot is black, while dynamic obstacles are red. The framework simultaneously estimates the ego pose and dynamic object motions, predicts the trajectories of moving obstacles, and plans a safe local path.}}
    \label{fig:front_graphic}
    \vspace{-6mm}
    % \vspace{\baselineskip}
\end{figure}

%To address the issue of operating in dynamic environments and to properly handle simultaneous optimization, we propose a unified framework, 

To address the challenges of operating in dynamic environments while properly managing simultaneous optimization, we propose \textit{DynoJEPP}, a unified framework for joint estimation, prediction and planning in dynamic environments.

Robust estimation in dynamic environments is achieved by leveraging recent advances in Dynamic SLAM~\cite{morris2024icra, morris2025dynosam, judd2024ijrr_mvo, bescos2021ral} that simultaneously localize the robot and map the scene while estimating the $\SE$ motions of dynamic objects.
%Due to the number of state variables and variety of representations possible in Dynamic SLAM~\cite{morris2024icra}, it is not always possible to extract the exact variables relevant for planning and prediction tasks from the complex graph.
%Therefore, DynoJEPP constructs a single factor graph by significantly expanding upon these Dynamic SLAM methods, and simultaneously estimates robot pose and object motions while predicting future object states and planning safe robot actions.
%An example of our framework operating in simulation is visualized in~\figref{fig:front_graphic}.

Due to the large number of state variables and variety of representations used in Dynamic SLAM~\cite{morris2024icra}, extracting the specific quantities required for prediction and planning is often non-trivial.
DynoJEPP addresses this by extending Dynamic SLAM into a single unified factor graph formulation that jointly estimates the robot pose and object motions, predicts future object states, and plans safe robot actions.
An example of the DynoJEPP framework operating in a simulation environment is shown in~\figref{fig:front_graphic}.

% Our approach utilises recent developments in Dynamic SLAM~\cite{morris2024icra, morris2025dynosam, judd2024ijrr_mvo, bescos2021ral} that allow simultaneous estimation of both the robot pose and motions of any observable dynamic objects 
%Therefore, our system simultaneously estimates the robot pose and dynamic object motions, predicts future states of moving obstacles and plans robot actions for safe navigation, an example of which is visualized in~\figref{fig:front_graphic}.
% To jointly localize dynamic objects and the robot while mapping the scene,
% many Dynamic SLAM systems~\cite{morris2025dynosam, morris2024icra, bescos2021ral, judd2024ijrr_mvo} have been proposed in recent years. 

% These systems construct complex factor graphs with a significant number of variables that all contribute to computing the current states of the robot and dynamic objects. 
% Such graphs cannot be easily compressed into light-weight states to be shared among estimation, prediction and planning modules. 
% % Additionally, incorporating past observations of dynamic objects improves the accuracy of trajectory prediction~\cite{huang2022survey}. 
% It is therefore even more essential to construct a joint factor graph to solve these problems together. 

To prevent planning and prediction from degrading estimation results in DynoJEPP, 
we further propose a novel \textit{directed factor} that
facilitates directional information flow between factor graph components.
We use this factor to propagate our estimation results directly into prediction and planning, whilst keeping estimation isolated from their influence.
To validate the directed factor's necessity and show the functionalities of our framework, 
we assess DynoJEPP on a variety of challenging scenarios in simulation. 
Our experiment results demonstrate that DynoJEPP is able to consistently navigate safely through dynamic environments.
By comparing our method with and without directed factors, we clearly show the necessity of controlling the influence of prediction, planning and estimation on each other to ensure successful operations in dynamic environments.

% Our proposal of directed factor
% In this paper, 
% we therefore propose a unified framework that jointly optimizes for prediction in addition to estimation and planning, enabling spatiotemporal planning in dynamic environments.
% Our system conducts estimation and prediction in motion space so as to not rely on any prior knowledge of object category, shape or motion model. 
% We further present a novel directed factor that protects the belief constructed from measurements and observations while allowing information from estimation to propagate directly into prediction and planning. 
% To validate our proposed framework, 
% we integrate it with a state-of-the-art Dynamic SLAM system and assess it with a wide variety of challenging scenarios in simulation. 
% Our experiment results demonstrate that this system is capable of both following and avoiding dynamic entities in the environment. 

To this end, the contributions of our work are as follows:
\begin{itemize}
    \item \textit{DynoJEPP}, a framework that jointly formulates and simultaneously optimizes robot and object state estimation, trajectory prediction and local planning in dynamic environments. To the best of our knowledge, this is the first factor-graph-based framework that jointly estimates dynamic object motion, predicts their evolution, and incorporates these predictions into robot control actions.
    % Mik: Removed "future" from "predicts their future evolution" to save space. 'evolution' already means future.

    \item A novel \textit{directed factor} that specifies the influence direction in a factor graph, allowing information flow among estimation, prediction and planning to be controlled.

    % \item \New{An extension of our formulation that jointly formulates planned ego trajectories and cooperative dynamic-object behavior, where the level of cooperation is parameterized within the simultaneously optimized problem and realized via directed factors.}
    % \item An extension of our formulation in which the level of dynamic object cooperation is parameterized within the simultaneously optimized problem and realized via directed factors.
    \item An extension of our formulation parameterizing the level of dynamic object cooperation within the simultaneous optimization, enabled via directed factors.
\end{itemize}

% 1. Joint system
% 2. Method - Directed factor
% 3. Predictions in motions space
% \begin{itemize}

%     \item A method for connecting the measurement-factor part of the graph with the cost-factor portion, while preserving accurate pose covariance estimated from measurements.

%     \item A trajectory prediction formulation in $\mathrm{SE}(3)$, in motion space, leveraging observed object motion and a static environment map.
    
%     \item Experimental Validation of the proposed approach in a jointly optimized system integrating dynamic SLAM, trajectory prediction, and local planning.
% \end{itemize}

% \textbf{---}

% Until now, existing works use ground truth knowledge on object motion - we can estimate them, very accurately. 
% Based on these estimations, we can do very good predictions. 
% Uncertainty of the estimation is meaningful and beneficial. 
% Better than purely reactive, and than conservative. 

% Note: we are not considering all the factors in prediction. 
% The main scope is that 
% 1. a good motion estimation is good for planning--planning result is not worse than using ground truth
% 2. prediction with uncertainty is helpful/useful for planning--not worse than using ground truth
% 3. The planning is good compared to a baseline method

\section{Related Work}
This section discusses recent literature on local planning and obstacle avoidance in dynamic environments, especially factor-graph-based simultaneous estimation and planning.
These are the most relevant fields to our work.

% the benefit of understanding environment dynamics to planning
% reactive vs predictive
% known motion vs estimated motion
% To navigate dynamic environments, 
% many works take on a two-stage structure. 
% For example, Bazzana~\etal\cite{Bazzana2023RAL} use a classic Dijkstra's algorithm for global planning, 
% and use Model Predictive Control (MPC) to avoid collisions with obstacles locally. 
% In such designs, the global plan provides a general guidance for the robot and does not change frequently, while the local planner responds to the constantly changing dynamic scene at a higher frequency. 

Predicting motions of dynamic objects in the scene has been shown to significantly improve the safety of obstacle avoidance~\cite{huang2022survey, lindqvist2020ral} and the quality of the planned route, e.g. less aggressive maneuvers and better smoothness~\cite{Finean2021icaps, Finean2023ras, King2022ICRA_SCATE} when compared to purely reactive replanning. 
Even assuming a simple constant velocity model can improve obstacle avoidance over short time horizons~\cite{lindqvist2020ral}. 
Nevertheless, dynamic objects in the real world often present complex motions, 
and assuming stringent motion models can limit the accuracy and safety of planning~\cite{Espinoza2023arxiv_STEAM, Finean2023ras}.
Frameworks such as SCATE~\cite{King2022ICRA_SCATE} incorporate predictions as fixed inputs rather than optimization variables, requiring perfectly known trajectories that are difficult to obtain in the real world.
In recent years, state-of-the-art Dynamic SLAM frameworks~\cite{morris2024icra, morris2025dynosam, judd2024ijrr_mvo} have achieved highly accurate motion estimations agnostic to object semantics. 
We therefore leverage such Dynamic SLAM framework in this work to enable accurate object motion predictions without prior knowledge, to better facilitate real-world deployments.

% Modeling uncertainties of the localization and mapping results has additionally been shown to improve the navigation safety~\cite{Nakamura2023icra} and produce smooth and natural trajectories~\cite{Indelman2015IJRR_GBS}.
% Nakamura and Bansal~\cite{Nakamura2023icra} demonstrate that modelling uncertainties from localisation and mapping improves the safety of planning for the real-world deployment. 
% Therefore, Agha-mohammadi~\etal\cite{Agha2018tro_SLAP} combines online planning and replanning under uncertainty in changing environments in Simultaneous Localization And Planning (SLAP). 
% Similarly, Ta~\etal\cite{Ta2014ICUAS_MPC} combines state estimation and MPC using factor graph.

% Joint localisation and planning
While the estimation, prediction and planning tasks can be addressed by individual decoupled components~\cite{Finean2023ras}, 
solving them simultaneously reduces redundancy as these are all trajectory optimization problems~\cite{mukadam2019ar_STEAP}, allowing necessary information to flow directly among them.
Robot state estimation and planning can be combined together by adapting Model Predictive Control (MPC) to a factor graph formulation~\cite{Bazzana2023RAL, Ta2014ICUAS_MPC}.
Bazzana~\etal\cite{Bazzana2023RAL} constructed two factor graphs that solve localization and planning successively. 
The planner is formulated as an MPC that receives the latest robot pose from the localization component. The authors compare their approach against the standard ROS navigation stack, demonstrating the suitability of solving MPC via maximum likelihood over a factor graph.
Ta~\etal\cite{Ta2014ICUAS_MPC} on the other hand connects state estimation and control into a jointly formulated factor graph.
% , connecting two components via current \Verify{and planned} optimal estimates of static landmarks and robot state.
% Dynamic obstacles naturally lead to incomplete or uncertain knowledge of the environment, and consequently call for planning under uncertainty.
% Modeling uncertainties of the localization and mapping results has additionally been shown to improve the navigation safety~\cite{Nakamura2023icra} and produce smooth and natural trajectories~\cite{Indelman2015IJRR_GBS}.
% % Nakamura and Bansal~\cite{Nakamura2023icra} demonstrate that modelling uncertainties from localisation and mapping improves the safety of planning for the real-world deployment. 
% Therefore, Agha-mohammadi~\etal\cite{Agha2018tro_SLAP} unify online planning and replanning under uncertainty in changing environments in Simultaneous Localization And Planning (SLAP). 
% Similarly, Ta~\etal\cite{Ta2014ICUAS_MPC} combines state estimation and MPC using factor graph.
Mukadam~\etal\cite{mukadam2019ar_STEAP} proposed Simultaneous Trajectory Estimation and Planning (STEAP),
which builds on the Gaussian Process Motion Planner (GPMP2)~\cite{Dong2016RSS_GPMP2, Mukadam2018gpmp2} and 
the Simultaneous Trajectory Estimation and Mapping (STEAM)~\cite{Anderson2015IROS_STEAM, Barfoot2014rss}. The method leverages Gaussian Process (GP) factors within a factor graph to enable continuous-time trajectory estimation and planning.

\begin{table}[t]
\footnotesize
\centering
\setlength{\tabcolsep}{2.8pt}
\caption{\small{Overview of factor-graph-based simultaneously optimized works. 
\textbf{Estimation} refers to trajectory estimation of the ego robot and dynamic objects.
\textbf{Prediction} refers to predicting dynamic object motions.
\textbf{Planning} refers to planning future states of the ego robot, either locally or globally.}
 $^\star$ -  SCATE assumes that the predictions are known and are part of a precomputed SDF.}
\label{tab:related_joint_work}
\begin{tabular}{c|cccc}
\toprule
 & \multicolumn{2}{c}{\textbf{Estimation}} & \textbf{Prediction} & \textbf{Planning} \\
 \textbf{Method} & \textit{Ego} & \textit{Dynamic Objects} &  &  \\
\midrule
\midrule
 STEAM \cite{Barfoot2014rss, Anderson2015IROS_STEAM} & \cmark & \xmark  & \xmark & \xmark\\
 \midrule
 STEAP \cite{mukadam2017rss_STEAP, mukadam2019ar_STEAP} & \cmark & \xmark & \xmark & \cmark \\
 \midrule
 SCATE \cite{King2022ICRA_SCATE} & \cmark & \xmark & \xmark \rlap{$^\star$}  & \cmark \\
 \midrule
 Ours & \cmark & \cmark & \cmark & \cmark  \\
\bottomrule
\end{tabular}
\vspace{-4mm}
\end{table}

However,~\tabref{tab:related_joint_work} shows that most of these works address static scenes~\cite{Barfoot2014rss, Anderson2015IROS_STEAM, mukadam2017rss_STEAP, mukadam2019ar_STEAP}, or mostly static scenes~\cite{King2022ICRA_SCATE}.
% with limited dynamics such as opening and closing office doors~\cite{Agha2018tro_SLAP}. 
Simultaneous Control and Trajectory Estimation (SCATE)~\cite{King2022ICRA_SCATE} jointly formulates estimation, planning and control within a unified factor graph to achieve collision-free navigation in environments with moving objects.
The predictive SCATE assumes perfect knowledge of dynamic obstacles, whereas reactive SCATE treats dynamic environments as static.
In their experiments, SCATE computes the robot state separately from control and planning. 
Contrastingly, our approach fuses Dynamic SLAM based estimation with object trajectory prediction, planning and control within a single factor graph.
% \Mik{SCATE is not mostly static. It considers either known, perfect predictions or is reactive. (They have predictive and reactive version). so they techincally "Estimate Dynamic objects" -\tabref{tab:related_joint_work}}
% We are instead fusing Dynamic SLAM with object trajectory prediction and MPC in one monolithic factor graph, to achieve local collision avoidance with dynamic obstacles.

% The development of joint optimization techniques using factor graphs, spanning from 2014 to present, is presented in \tabref{tab:related_joint_work}, where each entry reflects the specific set of components addressed by the corresponding work.

% \section{Method}

\begin{figure*}[t]    
    \centering
  \includegraphics[width=0.95\textwidth]{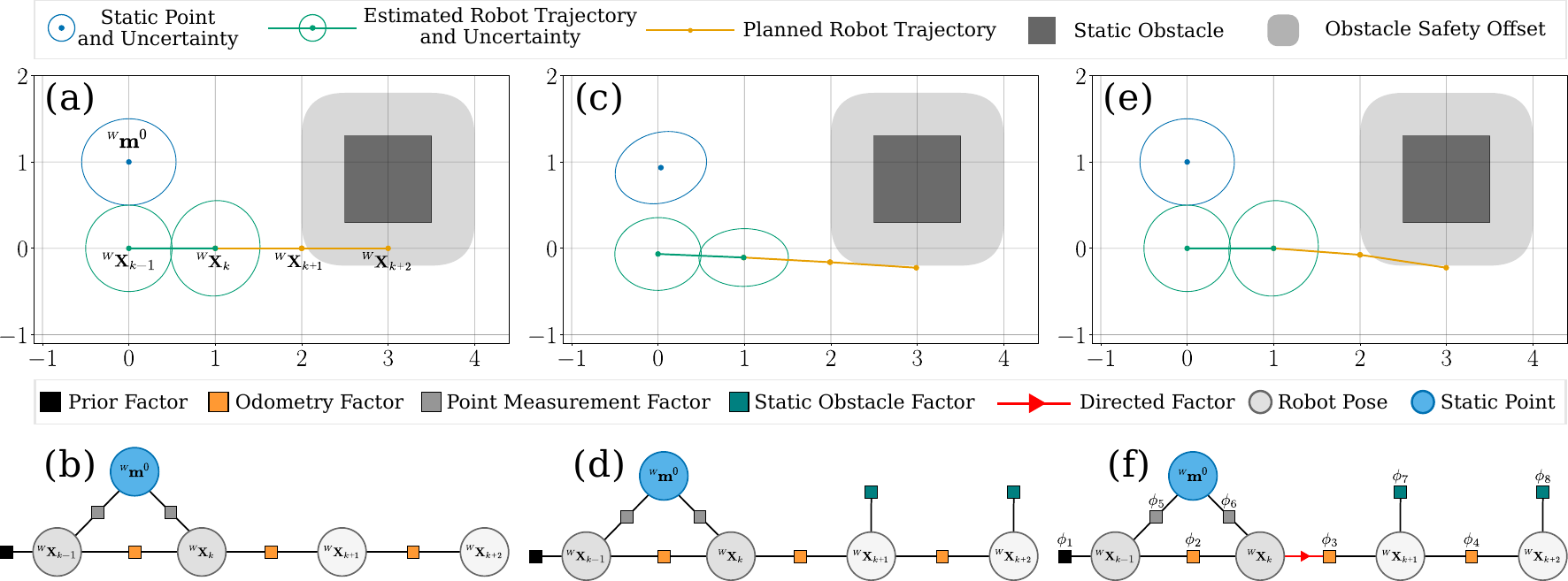}
   \caption{\small{A static 2D toy example demonstrating directed factor importance. 
   \textbf{(a, b)}: A robot observes a static point $\mpoint{\worldf}{}^0$ at poses $\campose{\worldf}{k-1}$ and $\campose{\worldf}{k}$. Future states $\campose{\worldf}{k+1}$ and $\campose{\worldf}{k+2}$ are planned without static obstacle factors. 
   \textbf{(c, d)}: Under bidirectional information flow, the introduction of static obstacle factors erroneously alters the estimated trajectory. 
   \textbf{(e, f)}: The proposed directed factor prevents planning from influencing state estimates while planning around the static obstacle.}}
    \label{fig:toy_example/plots}
    \vspace{-4mm}
    % \vspace{\baselineskip}
\end{figure*}
\section{Preliminaries}
\subsection{Notation}
\label{sec:notation}

% We define robot poses $\mathcal{X}$, velocities $\mathcal{V}$ and accelerations $\mathcal{A}$:
Our formulation defines robot poses, velocities and accelerations as:
\begin{equation*}
   \mathcal{X} = \{ \campose{\worldf}{k} \}_{{k \in \mathcal{K}}}, \; \mathcal{V} = \{ \vel{\camf_{k}}{k} \}_{k \in \mathcal{K}}, \; \mathcal{A} = \{ \acc{\camf_{k}}{k} \}_{k \in \mathcal{K}}
\end{equation*}
where $\{\worldf\}$ is the world frame and $\campose{\worldf}{k} \in \SE$ is the robot's body-fixed frame. 
$\mathcal{K}$ is the set of all time-steps. 
As the time-step of the velocity and acceleration is always identical to that of their reference frames, we simplify their notations to $\vel{\camf_{k}}{k} \equiv \vel{\camf}{k}, \acc{\camf_{k}}{k} \equiv \acc{\camf}{k}$.

In a dynamic scene, 
an object pose at time-step $k$ in frame $\{\worldf\}$ is denoted as $\objpose{\worldf}{k} \in \SE$ and is associated with a body-fixed reference frame $\{\objf_k\}$. 
A 3D point is denoted as $\mpoint{}{}^{i}$, where $i$ 
denotes the point index.
% additionally indicates correspondences.
% The homogeneous coordinate of any point $\tilde{\mathbf{m}}^i\in\mathbb{R}^3$ is $\mpoint{}{}^{i} = \left[\tilde{\mathbf{m}}^i, 1\right]^\top$, where $i$ additionally indicates correspondences.

% \begin{equation*}
%    \mathcal{X} = \{ \campose{\worldf}{k} \}_{{k \in \mathcal{K}}}, \quad \mathcal{L} = \{ \objpose{\worldf}{k}^j \}{\substack{{j \in \mathcal{J}_k} \\ {k \in \mathcal{K}_j}}}
% \end{equation*}

% A 3D point in the world frame at time-step $k$ is denoted as $\mpoint{\worldf}{k}^i$, where $i$ indicates correspondences across frames. 
% To minimize notation clutter, we liberally omit indices $i$ and $j$ when there is no ambiguity.
% For instance, when a point is time-invariant in its reference frame, e.g. a static point $\mpoint{\worldf}{}^{i}$ in the world frame, we omit its time-step $k$ in the notation. 

We denote the rigid-body motion of an object between  arbitrary time-steps $A$ and $B$ as $\objmotion{\worldf}{A}{B} \in \SE$ following the Dynamic SLAM works~\cite{Henein20icra, morris2025dynosam}:
\begin{equation}
    \objpose{\worldf}{B} = \objmotion{\worldf}{A}{B} \: \objpose{\worldf}{A}\text{.}
\label{equ:object_motion_from_pose_def}
\end{equation}
% By expressing this motion in $\{\worldf\}$, all points on the rigid-body object $j$ can be transported from one time-step to another using a common $\SE$ transformation~\cite{morris2025dynosam, zhang2020vdoslam, Henein20icra}:
% \begin{equation}
%     \mpoint{\worldf}{B} = \objmotion{\worldf}{A}{B} \: \mpoint{\worldf}{A}\text{.}
%     \label{equ:general_point_motion}
% \end{equation}
% Following the Dynamic SLAM works~\cite{morris2025dynosam}, we assume objects are rigid and by parameterizing the motion in the world frame, the motion becomes \textit{independent} of the object reference frame $\{\objf\}$, allowing $\{\objf\}$ to be placed arbitrarily on the object.

% As the proposed system is concerned of planning and controlling the motions of a robot, we also plan for the robot's velocity $\vel{\camf_{k}}{k}$ and acceleration $\acc{\camf_{k}}{k}$ at each time-step. 
% We further parameterize the robot's velocity $\vel{\camf_{k}}{k}$ and acceleration $\acc{\camf_{k}}{k}$ in our system for planning and controlling the robot's motion. 
% As the time-step of the velocity and acceleration is always identical to that of their reference frames, we simplify their notations as follows: 
% \begin{equation*}
%     \vel{\camf_{k}}{k} \equiv \vel{\camf}{k}, \quad \acc{\camf_{k}}{k} \equiv \acc{\camf}{k}\text{.}
% \end{equation*}

\subsection{Factor Graph Optimization}

%%%%%%%%%%% - Old version start
% Factor graphs are a family of graphical models that enable probabilistic inference over a set of state variables $\theta$ given a set of factors $\phi$, which encode relationships between subsets of variables.
% Each factor $\phi_i$ in the graph is of the form:
% \begin{equation}
%     \phi(\cdot)_i \propto \text{exp} \Bigl\{ -\frac{1}{2} \factor{\mathbf{r_i}}{} \Bigr \}\text{,}
% \label{equ:factor_form}
% \end{equation}
% where $\mathbf{r}(\cdot)$ is a residual error function and $\Sigma$ is the associated covariance matrix.
% Assuming each $\phi_i$ represents a likelihood over the set of involved variables corrupted by zero-mean Gaussian noise, maximum a posteriori (MAP) inference in the factor graph is equivalent to minimizing the sum of squared Mahalanobis
% residuals, yielding the nonlinear least-squares problem:
% \begin{equation}
%      \mathbf{\theta}^{\text{MAP}} =\underset{\mathcal{\theta}}{\mathrm{argmin}} \sum_i \factor{\mathbf{r_i}}{} \text{.}
%      \label{equ:nonlinearLS}
% \end{equation}
% The nonlinear least-squares problem is solved iteratively.
% At each iteration, the residuals in~\eqref{equ:nonlinearLS} are linearized around the current estimate $\theta$, yielding the normal equation:
% \begin{equation}
% \mathbf{J}^\top \mathbf{J} \, \boldsymbol{\delta} = - \mathbf{J}^\top \mathbf{r},
% \label{equ:normal_equationLLS}
% \end{equation}
% where $\mathbf{J}$ is the Jacobian of the stacked residuals $\mathbf{r}$ evaluated at the current estimate $\theta$.
%%%%%%%%%%% - old version above

%%%%%%%%%%%%%%%
A factor graph represents a function over a set of variables $\boldsymbol{\theta}$ as a product of factors where each factor $\phi_i$ encodes a relationship over a subset of variables expressing how those variables are coupled, e.g. through measurements, costs, or constraints and is  written in exponential residual form:
\begin{equation}
\phi_i(\theta_i)
\propto
\exp\!\left(
-\frac{1}{2}\,\|r_i(\cdot)\|_{\Sigma_i}^2
\right),
\label{equ:factor_form}
\end{equation}
where $r_i(\cdot)$ is a residual function and $\Sigma_i$ is a positive definite weighting matrix.
Maximum a posteriori (MAP) inference corresponds to minimizing the total objective:
\begin{equation}
\boldsymbol{\theta}^{\star}
=
\arg\min_{\boldsymbol{\theta}}
\sum_i
\|r_i(\cdot)\|_{\Sigma_i}^2 .
\label{equ:nonlinearLS}
\end{equation}
This nonlinear least–squares problem is solved iteratively by linearizing the residuals
around the current estimate of $\boldsymbol{\theta}$. Denoting the stacked residual vector by $\mathbf r$
and its Jacobian by $\mathbf J$, the increment $\boldsymbol{\delta}$ satisfies:
\begin{equation}
\mathbf J^\top \mathbf J\,\boldsymbol{\delta}
=
-\,\mathbf J^\top \mathbf r .
\label{equ:normal_equationLLS}
\end{equation}
%%%%%%%%%%%%%%%%%%%%%%%%%%
\section{Directed Factor}

\label{sec:directed}

% \begin{figure}[b]
%     \centering
%   \includegraphics[width=\columnwidth]{figs/toy_example/factor_graph.pdf}
%    \caption{\small{Factor Graph Example in which  variables $\campose{\worldf}{k-1}$, $\campose{\worldf}{k}$ and $\mpoint{\worldf}{}^0$ are variables in the estimation problem, and $\campose{\worldf}{k+1}$ and $\campose{\worldf}{k+2}$ are part of the planning problem. The planned camera poses are near the static obstacle, and static obstacle factors contribute non-zero errors.}}
%     \label{fig:toy_example/factor_graph}
%     % \vspace{-4mm}
%     % \vspace{\baselineskip}
% \end{figure}

% It is vital that future robot states should not influence current and past states because they are based on existing measurements. 
Future robot plans should not influence estimates of current or past states, as these are determined entirely by existing measurements.
However, existing frameworks that simultaneously solve estimation and planning~\cite{mukadam2017rss_STEAP, mukadam2019ar_STEAP, King2022ICRA_SCATE} allow information to propagate bidirectionally between past and future states.
In this section, we demonstrate that by allowing this, the planning will degrade the estimation results.

In~\figref{fig:toy_example/plots}, we construct a static 2D toy example highlighting the degradation effect and its resolution by the directed factor. 
The robot moves from $\campose{\worldf}{k-1}$ to $\campose{\worldf}{k}$ and plans two poses ahead using a constant velocity model as shown in~\figref{fig:toy_example/plots}(a).
For estimation, 
the robot measures a static map point $\mpoint{\worldf}{}^0$ at time-steps $k-1$ and $k$, seen in the corresponding factor graph~\figref{fig:toy_example/plots}(b).
In~\figref{fig:toy_example/plots}(c, d) a similar setup considers the static obstacle in the plan through the use of static obstacle factors which encourage collision-free trajectories.

In the initial setup (\figref{fig:toy_example/plots}(a, b)), the estimation correctly reflects all measurements, but the robot plans into unsafe regions.
By taking into account the static obstacle (\figref{fig:toy_example/plots}(c, d)), the planned trajectory becomes collision free; however, estimated states and their uncertainties are erroneously influenced by additional constraints added by factors involved in planning.
The goal is therefore to create a single factor-graph that retains the accurate estimation from~\figref{fig:toy_example/plots}(a) while also planning collision-free trajectories, as in~\figref{fig:toy_example/plots}(c).

To achieve this, we propose a directed factor that enforces unidirectional information flow, as illustrated in~\figref{fig:toy_example/plots}(f). 
The directed factor connects $\campose{\worldf}{k}$ and $\campose{\worldf}{k+1}$, creating a directional connection between estimation and planning. 
Past variables $\mpoint{\worldf}{}^0, \campose{\worldf}{k-1}$ and $\campose{\worldf}{k}$ become independent of planning, 
while planning remains dependent on the current Maximum a Posterior (MAP) estimate of $\campose{\worldf}{k}$. 
The estimated and planned trajectory from this new factor graph is visualized in~\figref{fig:toy_example/plots}(e). 
Compared with~\figref{fig:toy_example/plots}(a), this new factor graph yields equivalent estimation results and uncertainties while planning around the obstacle.

\begin{figure}[h]
    \centering
    % \vspace{-2mm}
    \includegraphics[width=\linewidth]{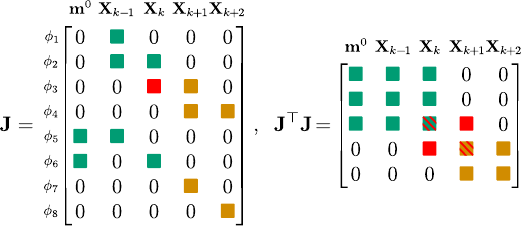}
   \caption{\small{Sparsity pattern of the factor graph in~\figref{fig:toy_example/plots}(f). Estimation-related entries are green, and planning-related entries are orange. Solid red entries are zeroed by the directed factor, while red-striped entries are partially affected but remain non-zero.}}
   \vspace{-2mm}
    \label{fig:directed_factor_toy_jacobians} 
\end{figure}

The directed factor is realized via \textbf{Jacobian manipulation}. 
By zeroing the Jacobian with respect to the source variable, we ensure it remains unaffected by the factor's gradient.
As illustrated in~\figref{fig:directed_factor_toy_jacobians}, this modification removes the non-zero entries that would otherwise form off-diagonal cross terms in $\mathbf{J}^\top \mathbf{J}$.
Based on~\eqref{equ:normal_equationLLS}, the resulting normal equation for the factor graph in~\figref{fig:toy_example/plots}(f) becomes:
\begin{equation}
\mathbf{J}^\top \mathbf{J} \boldsymbol{\delta} \!=\! 
\begin{bmatrix}
{\color{EgoEstimation}\blacksquare} \mathbf{m}^0 , \mathbf{X}_{k-1}, \mathbf{X}_{k} \\[0.3em]
{\color{EgoEstimation}\blacksquare} \mathbf{m}^0 , \mathbf{X}_{k-1}, \mathbf{X}_{k} \\[0.3em]
{\color{EgoEstimation}\blacksquare} \mathbf{m}^0 , \mathbf{X}_{k-1}, \mathbf{X}_{k} \\[0.3em]
{\color{EgoPlan}\blacksquare} \mathbf{X}_{k+1}, \mathbf{X}_{k+2} \\[0.3em]
{\color{EgoPlan}\blacksquare} \mathbf{X}_{k+1}, \mathbf{X}_{k+2}
\end{bmatrix}
\!=\!
\begin{bmatrix}
{\color{EgoEstimation}\blacksquare} \mathbf{r}_5, \mathbf{r}_6 \\[0.3em]
{\color{EgoEstimation}\blacksquare} \mathbf{r}_1, \mathbf{r}_2, \mathbf{r}_5 \\[0.3em]
{\color{EgoEstimation}\blacksquare} \mathbf{r}_2,\mathbf{r}_6 \\[0.3em]
{\color{EgoPlan}\blacksquare} \mathbf{r}_3, \mathbf{r}_4, \mathbf{r}_7 \\[0.3em]
{\color{EgoPlan}\blacksquare} \mathbf{r}_4, \mathbf{r}_8
\end{bmatrix} \!=\! -\mathbf{J}^\top \mathbf{r}\text{.}
\label{equ:toy_example_normal_equationLLS}
\end{equation}
The absence of cross terms reflects the structural decoupling between estimation and planning. Importantly, the RHS of~\eqref{equ:toy_example_normal_equationLLS} shows that the update for $\mathbf{X}_{k+1}$ still depends on the current linearized value of $\mathbf{X}_{k}$ through the residual $\mathbf{r}_3$, despite the zeroed cross terms.
Because cross terms are removed, uncertainty does not propagate from the directed variable. For the local planner, this is not restrictive, as the robot remains well localized relative to nearby objects through recent measurements. 
In practice, global uncertainty is accounted for by a separate module that provides the objective for the local goals.
Furthermore, since estimation, prediction and planning are distinct cost objectives, this decoupling ensures that weights or covariances in one module do not influence the other. 
In large dynamic graphs, this prevents the growing estimation sub-graph from shifting the overall objective balance. 
By isolating these cross-module interactions, the framework maintains consistent behavior without frequent retuning.

\begin{figure*}[t]
	\centering
    \includegraphics[width=\textwidth]{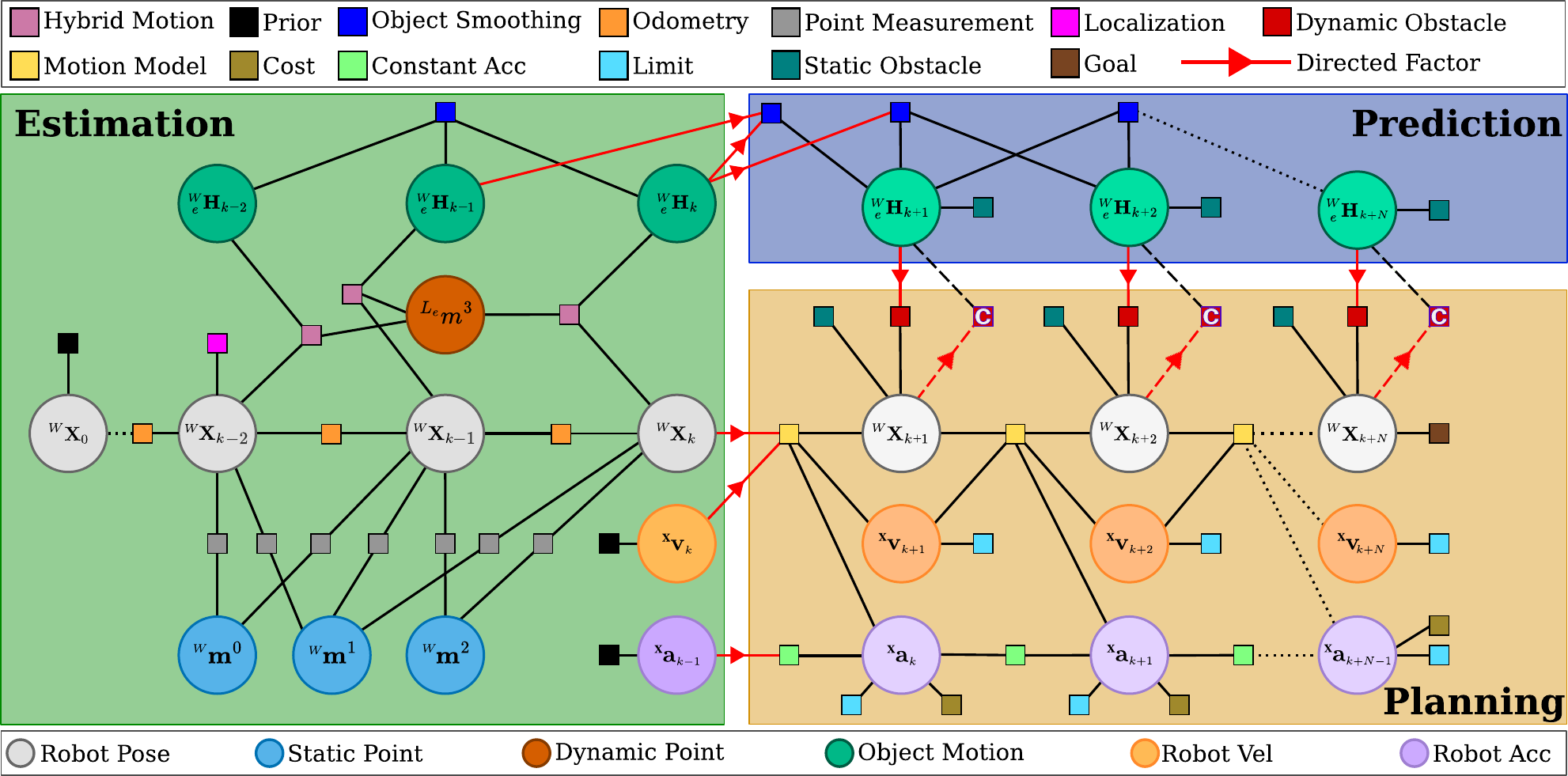}
	\caption{\small{Proposed DynoJEPP factor graph for jointly formulated estimation (green), prediction (blue), and planning (orange). Variables and factors are depicted as circular and square nodes, respectively. 
    Directed factors (red arrows) manage information flow between components to ensure safe operation in dynamic environments. Dynamic obstacle factors marked with `C' are exclusive to the C-DynoJEPP extension.}}
    \label{fig:system_factor_graph}
    \vspace{-4mm}
\end{figure*}

\section{DynoJEPP}
\label{sec:DynoJEPP} 
%DynoJEPP takes in RGB-D sensor data as input and generates a static-dynamic... 
DynoJEPP processes sensor measurements to generate a static-dynamic map estimate, predict object motions, and plan local trajectories and control actions.
Prediction and planning are scene-aware and are optimized simultaneously with the estimation as new data is received.
% Both planning and prediction are scene-aware and are optimized simultaneously with the estimation, given new sensor measurements of the environment.
% We plan directly in the action space (velocity) of the robot to ensure that the plan is feasible given physical limits. 

\subsection{Jointly Formulated Estimation, Prediction \& Planning}
DynoJEPP solves the optimization problem represented by the factor graph shown in \figref{fig:system_factor_graph}, which integrates estimation, prediction and planning. 
The estimation component resembles a typical SLAM factor graph, where $\phi$ in~\eqref{equ:factor_form} represents measurement models and $\theta$ in~\eqref{equ:nonlinearLS} denotes estimated variables. 
Our formulation additionally incorporates variables representing  object motion predictions, planned robot states, and control commands, with factors modeling the costs for desired object and ego trajectories.

% To ensure that our approach is both dynamic-object-aware and informed by estimation, the three components of the factor-graph are connected.
% However, as discussed in~\secref{sec:directed} this can result in degradation of the estimation and prediction causing unsafe plans.
%Given the joint structure, our formulation relies heavily on the proposed directed factors (highlighted by the red arrow in~\figref{fig:system_factor_graph}) to control information flow across components: the estimation module remains unaffected by prediction and planning, prediction depends only on estimation, and planning depends on both estimation and prediction. 

Given the joint formulation, our formulation critically depends on the proposed directed factors (highlighted by the red arrow in~\figref{fig:system_factor_graph}) to enforce a hierarchical coupling: estimation remains independent of prediction and planning, prediction depends on estimation, and planning is informed by both.
This architecture acts as a hybrid of two classical models: directed acyclic graphs for causal dependencies between modules, and undirected graphs for the internal structure of each component \cite{koller2009probabilistic}. By ensuring information only flows downstream, this structure prevents prediction and planning objectives from altering the state estimate.

\subsection{State Estimation}
\label{sec:estimation}

To accurately estimate for the dynamic scene, our method builds upon the Hybrid Dynamic SLAM formulation proposed in~\cite{Morris2025ral_dynosam}.
This approach estimates the robot pose $\campose{\worldf}{k}$ and the per-frame motion $\objmotion{\worldf}{e}{k}$ of each object.
The static structure and per-object map are additionally estimated for and refined over multiple time-steps.
Rather than directly estimating the object pose at each time-step $k$,
this formulation estimates the motion $\objmotion{\worldf}{e}{k}$ which maps any fixed frame on the object from time-step $e$ to $k$ following~\eqref{equ:object_motion_from_pose_def}. 
The object pose $\objpose{\worldf}{k}$ can be recovered for any $k$ given an initial frame $\{\objf_e\}$. All factors in the estimation component follow~\cite{Morris2025ral_dynosam}, and are omitted here due to space constraints, slight modifications are detailed below.

This work implicitly models all objects as spheres, therefore their pose is considered $\objpose{\worldf}{e} = \othpose{\worldf}{C}{e}$ where $\othpose{\worldf}{C}{e}$ is the object’s Center of Mass (CoM) at $e$.
Thus the estimated motion will track the object's CoM in the world frame for each $k$ as $\othpose{\worldf}{C}{e}$ is a constant for each object:
\begin{equation}
    \othpose{\worldf}{C}{k} = \objmotion{\worldf}{e}{k} \: \othpose{\worldf}{C}{e}\:.
\label{equ:com_k}
\end{equation}
In practice, $\othpose{\worldf}{C}{e}$ can be approximated from accumulated reconstructions, learning-based methods, or known models.
\\
We denote all residuals that depend on the object pose with $\mathbf{r}(\othpose{\worldf}{C}{k})$.
This indicates that they are connected to the motion variable $\objmotion{\worldf}{e}{k}$ and include the constant $\othpose{\worldf}{C}{e}$ such that the full pose can be recovered via~\eqref{equ:com_k}.

At each step $k$, observations of both the static and dynamic environment are received and integrated using \textit{point measurement} and \textit{hybrid motion factors}.
Point measurement factors link the robot pose to tracked static points $\mpoint{\worldf}{}$, while each hybrid motion factor connects observations of tracked dynamic points to the rigid-body motion model $\objmotion{\worldf}{}{}$ and the corresponding robot pose. 
The \emph{odometry factor} links consecutive robot poses through a measured relative transformation, obtained from visual odometry or, when available, from the IMU.

Finally, the \textit{object smoothing factor} imposes a constant motion model between pairs of motion variables, promoting locally consistent and physically plausible trajectories:
\begin{equation}
     \mathbf{r}_{\delta H} =  \log  \Bigg[\left( \othpose{\worldf}{C}{k-2}^{-1} \:   \othpose{\worldf}{C}{k-1}\right)^{-1}
   \left(\othpose{\worldf}{C}{k-1} ^{-1} \: \othpose{\worldf}{C}{k} \right)  \Bigg] ^\vee\text{.}
\label{equ:object_smoothing_factor}
\end{equation}
As described in~\cite{Morris2025ral_dynosam} this factor is conceptually similar to constant velocity model but encoded in $\SE$.

\subsection{Object Motion Prediction}
\label{sec:prediction}

% Variables
% Factors
% Directed Factors

The prediction module in~\figref{fig:system_factor_graph} utilizes the estimated $\SE$ motions of dynamic objects from~\secref{sec:estimation}, to forecast future trajectories over a horizon of $N$ steps.
\\
Since our method builds upon the Dynamic SLAM formulation proposed in~\cite{morris2025dynosam, Morris2025ral_dynosam}, the estimated object motions are inherently smooth and require no additional preprocessing~\cite{kliniewski2025}, enabling future motions $\mathcal{H}_{k+1:k+N}$ to be predicted using only the two most recent object motion estimations. 
This is implemented via the \textit{object smoothing factor}~\eqref{equ:object_smoothing_factor}, which enforces a constant velocity motion model and maintains temporal consistency.
% Similarly, the estimator enforces temporal consistency, making the full motion history unnecessary for predictions. \comments{estimator? What's similar in this?}
In order to prevent future predictions from influencing past estimates, we use directed factors whenever estimation variables $\objmotion{\worldf}{e}{k-1}$ or $\objmotion{\worldf}{e}{k}$ are involved.

\subsection{Local Planning in Action Space}
\label{sec:planning}

% \Mik{Need something like: "Estimation and prediction are in $\mathrm{SE}(3)$, 
% while control is executed in $\mathrm{SE}(2)$ under a planar motion assumption, 
% as the robot and obstacles in our experiments are all grounded."}

% \Mik{Need to mention specifically that v is linear and angular velocity}
% Variables
% Factors
% Directed Factors

% In this case the dynamic obstacle factor~\eqref{eq:dynamic_obstacle_factor} is used as the mission factor and a goal factor~\eqref{equ:goal_factor} is added to the $k+N^{\text{th}}$ camera pose with mean equal to the calculated carrot point

%  When in object following mode, the robot will seek to remain a predefined distance $d_{\text{ros}}$ away from the object's CoM and maintain some relative heading $\theta_{\text{ros}}$. For all following experiments $d_{\text{ros}}=2$ \Yiduo{unit? \SI{2}{\meter}?} and $\theta_{\text{ros}} = 0$.

% Here, the target follow factor~\eqref{equ:follow_factor} is the mission factor and no goal factor is required. 
% The robot will continue to follow the object while in view, or use the last set of $N$ predictions to continue following.

% \textbf{-------}

% Intro - the third component
The planning component of our framework (\figref{fig:system_factor_graph}) employs a factor-graph-based MPC to optimize a local trajectory over a horizon of $N$ steps. 
The optimized variables are future robot poses $\mathcal{X}_{k+1:k+N}$, velocities $\mathcal{V}_{k+1:k+N}$, and accelerations $\mathcal{A}_{k:k+N-1}$.
While the estimation and prediction modules operate in $\mathrm{SE}(3)$, control in our experiments is executed in $\mathrm{SE}(2)$ under a planar motion assumption.
The formulation can naturally extend to $\mathrm{SE}(3)$.
All factors that involve the planned robot pose follow the mapping: $\campose{}{k} \in SE(3) \mapsto \mathbf{x}_k = (x_k, y_k, \theta_k) \in \mathrm{SE}(2)$.

We follow the work of Bazzana~\etal\cite{Bazzana2023RAL} and model our robot as a unicycle controlled in translational and angular velocities such that $\mathbf{v}_k = [ v_{k}, \omega_{k} ]$
using the kinematic model $\mathbf{x}_{k+1} = f(\mathbf{x}_k, \mathbf{v}_k)$:
\begin{equation}
\begin{split}
   x_{k+1} &= x_k + v_{k} \Delta t \cos\left(\theta_k + {\omega_k \Delta t}/2 \right) \\
y_{k+1} &= y_k + v_{k} \Delta t \sin\left(\theta_k + {\omega_k \Delta t}/2\right) \\
\theta_{k+1} &= \theta_k + \omega_{k} \Delta t 
\label{eq:kinematics}
\end{split}
\end{equation}
which propagates the current $\mathrm{SE}(2)$ pose $\mathbf{x}_{k}$ to the next pose $\mathbf{x}_{k+1} $ over an integration interval $\Delta t$.

To inform future velocities and accelerations in the MPC, the robot requires information about its current state; 
thus, the known current velocity $\mathbf{v}_{k}$ and previously applied acceleration $\mathbf{a}_{k-1}$ are fixed via a \textit{prior factor}.
Furthermore, all future-related factors connected to known or estimated variables, i.e. $\campose{\worldf}{k}$, $\mathbf{v}_{k}$, $\mathbf{a}_{k-1}$ are directed to prevent planned states from influencing the estimation.

A \textit{motion model factor} models the kinematics in~\eqref{eq:kinematics} to ensure consistency with the robot's dynamics, relating acceleration to velocity and pose updates:
\begin{equation}
    \mathbf{r}_{\text{dyn}} = 
        \begin{bmatrix}
        \mathbf{X}_{k+1}^{-1} \: f(\mathbf{X}_{k}, \vel{}{k+1}) \\
        \vel{}{k+1} - (\vel{}{k} + \acc{}{k} \Delta t)  
    \end{bmatrix}\text{.}
\end{equation}
We utilize \textit{limit factors}~\cite{King2022ICRA_SCATE} to keep velocities $\mathcal{V}_{k+1:k+N}$ and accelerations $\mathcal{A}_{k:k+N-1}$ within desired bounds.
Unary \textit{cost factors} $\mathbf{r}_c = \mathbf{a}_k$ penalize non-zero accelerations to encourage constant velocity, 
while \textit{constant acceleration factors} $\mathbf{r}_{ca} = a_k - a_{k-1}$ smooth control inputs.
% Up till now we have presented a generic MPC structure that can be solved with a factor graph.
Finally, a \textit{goal factor} $r_{\text{goal}} = \campose{}{\text{goal}}^{-1} \: \campose{}{k+N}$ is applied to the $(k+N)^{\text{th}}$ pose to guide the robot toward a local goal.
This mission objective can be replaced or augmented by other factors attached to the final or all planned poses, depending on the desired task.

\subsection{Static Scene-Aware Prediction \& Planning}
\label{sec:static_scene_aware}

% Variables
% Factors
% Directed Factors

% \Mik{Structure should be: Intro as it is. Static obstacle factor for motion prediction. 
% The changing role of object smoothing factor when we include the static obstacle prediction. 
% Static obstacle factor for planning. The need for additional localization factor when we use the precomputed map that we need to localize against to query correct points.}

% While online dense reconstruction is outside the scope of this work, it can be performed via existing methods~\cite{Oleynikova2017iros_voxblox, Lan2025ral_vdbgpdf, Wang2025icra_dynorecon}.
% \comments{I don't think this sentence belongs here. Citations are not necessary.}

To incorporate static scene awareness, the framework interfaces with a static scene map via a Euclidean Signed Distance Field (ESDF)~\cite{Oleynikova2017iros_voxblox}. This information is integrated into the formulation through \textit{static obstacle factors}:
\begin{equation}
    \mathbf{r}_{\text{sdf}} =
    \begin{cases}
    d_{\text{os}}-\text{esdf}(\othpose{\worldf}{P}{k}), & \text{esdf}(\othpose{\worldf}{P}{k}) < d_{\text{os}} \\
    0, & \text{otherwise}
    \end{cases}\text{,}
    \label{eq:static_obstacle_factor}
\end{equation}
% where $d_\text{os}$ accounts for the radius of the \Verify{robot or object} plus a safety offset.
% Depending on its use in the prediction or planning component, 
where the evaluated pose $\othpose{\worldf}{P}{k} \in \{\othpose{\worldf}{C}{k}, \campose{\worldf}{k}\}$ corresponds to either a predicted object pose or a planned robot pose. Notably, $\othpose{\worldf}{C}{k}$ is a function of the predicted motion $\objmotion{\worldf}{e}{k}$ following~\eqref{equ:com_k}.
The function $\text{esdf}(\othpose{\worldf}{P}{k})$ looks up the distance value in the static map using coordinates given by the translational component of $\othpose{\worldf}{P}{k}$, while $d_\text{os}$ accounts for the respective robot or object radius plus a safety offset.

In prediction, these factors relax the constant-motion assumption to yield more realistic behaviors near obstacles, while \textit{object smoothing factors} ensure the resulting motions remain smooth.
In planning, they provide the necessary information for collision-free navigation.

Since the ESDF map used by these factors is precomputed, state estimation is supported by low-frequency global localization factors, to maintain alignment with the static environment. In practice, this can be achieved by online mapping~\cite{reijgwart2020ral_voxgraph}.

\subsection{Dynamic Scene-Aware Prediction \& Planning}
% Dynamic obstacle part
To incorporate dynamic scene awareness, the formulation links predicted object motions with planned robot states via directed \textit{dynamic obstacle factors}, which penalize proximity to moving objects: 
\begin{equation}
\begin{aligned}
    \epsilon & = d_{\text{ros}} - \text{range}( \campose{\worldf}{k}, \: \othpose{\worldf}{C}{k})\text{,}\\
    \mathbf{r}_{\text{obs}} &=
    \begin{cases}
     \epsilon, & \epsilon  > 0 \\
    0, & \text{otherwise}
    \end{cases}\text{,}
\end{aligned}
    \label{eq:dynamic_obstacle_factor}
\end{equation}
where $d_{\text{ros}}$ is the sum of the robot and object radii plus a safety offset.
While object radii can be obtained from dynamic mapping~\cite{Wang2025icra_dynorecon}, they are assumed known in this work.
% radii is plural of radius in American English
These factors are directed to ensure the robot plan adapts to predictions without affecting the predicted object motions.

% \begin{figure}
%     \centering
%     \includegraphics[width=0.95\columnwidth]{figs/architecture_dependability.pdf}
%     \caption{\Verify{Directed factor architecture in DynoJEPP vs Cooperative DynoJEPP.}}
%     \vspace{-4mm}
%     \label{fig:dyno_jepp_vs_cooperative}
% \end{figure}
 
% Cooperative DynoJEPP (C-DynoJEPP) extends the DynoJEPP framework, by incorporating cooperative behaviors of predicted dynamic objects into the simultaneous optimization, enabling less conservative local plans. This is achieved by employing directed \textit{dynamic obstacle factors} from planning to prediction, the symmetric counterpart to~\eqref{eq:dynamic_obstacle_factor}, as indicated by the `C' factors in~\figref{fig:system_factor_graph}.
Cooperative DynoJEPP (C-DynoJEPP) extends the DynoJEPP framework by incorporating cooperative behaviors of predicted dynamic objects into the simultaneous optimization, enabling less conservative local plans.
Cooperative behavior assumes mutual influence: the ego robot reacts to the object's motion, while the dynamic object is expected to react to the robot's actions.
We model the interaction between the robot’s future plan and the object's predicted motion using the \textit{dynamic obstacle factors} in~\eqref{eq:dynamic_obstacle_factor}. As shown by the 'C' factors in \figref{fig:system_factor_graph}, these are now directed from planning to prediction to explicitly model the influence of the robot's intended path on the object prediction.

C-DynoJEPP introduces a directed cycle between prediction and planning, enabling mutual influence while maintaining estimation independent.
% This simultaneous influence is inefficient to replicate with a decoupled architecture, which would require alternating between modules and fixing intermediate solutions. 
% Furthermore, standard undirected factors are insufficient here, as they impose a uniform weighting that eliminates the flexibility to independently tune prediction and planning costs.
Standard undirected factors are insufficient here, as their uniform weighting eliminates the flexibility to independently tune prediction and planning costs. 
Alternatively, replicating this simultaneous influence with a decoupled architecture is inefficient, requiring alternating between modules and fixing intermediate solutions. 
% While module weights are fixed in this work, in practice, the cooperativeness of predicted objects can be dynamically adjusted during execution to reflect updated environmental context.
The degree of cooperation is governed by the `C' factor weights. 
While these remain constant throughout the execution in this work, updating them online would allow shifting between conservative and assertive behavior of agents.

% While C-DynoJEPP weights are fixed in this work, the degree of predicted object cooperativeness can be adjusted online by changing the `C' factor weights as the object's behavior is observed. 
% This allows the framework to shift between conservative and cooperative assumptions as confidence in an agent's intent changes.

% Mention architecture, additional factor, fact that this is not possible via decoupling. Purpose is to show the versatility in the use of directed factor.

% \begin{figure}[!h]
%     \centering
%     \includegraphics[width=0.95\columnwidth]{figs/C_DynoJEPP_factors.pdf}
%     \caption{C-DynoJEPP}
%     \hfill
%     % \label{fig:}
% \end{figure}

% Mik here
\section{Experiments}

% \comments{\textbf{\underline{This is wrong!} We don't know if initialization is the only reason, so we should be more general and focus on the estimation.} \Mik{I input the text at the end of \Verify{DynoJEPP - Navigation.} We may still need to explain the initialization somewhere earlier tho.} \textbf{The Problem:} By simultaneously optimizing estimation, prediction and planning, the plan and prediction optimization process differs from the decoupled as the current available robot pose and object motion change with every iteration. This leads to different initialization and executing different control commands. As can be seen in~\tabref{tab:experiments_quantitative} the dynamic object estimation between these two configurations is similar across all the runs. The number of steps differs as for instance in experiment 10, the directed method lead to robot waiting for the dynamic object to pass whereas decoupled method decided to cut through and go around. (If we run the same experiment several times, they could make different decisions.) Evaluating initialization is outside of the scope of this work and can be considered a design choice (starting from a previously optimized solution or creating a fresh one). The important part is that they both maintain the correct estimation and don't crash.}

This section evaluates the impact of directed factors within the DynoJEPP and C-DynoJEPP frameworks.
Rather than comparing to external baselines, we analyze how these factors regulate inter-module information flow
% the interaction between estimation, prediction, and planning 
to prevent estimation degradation and enable safe planning.
We compare three configurations: \textit{Undirected}, allowing free bidirectional information propagation between modules;
\textit{Directed}, which regulates flow via directed factors; and \textit{Decoupled}, a baseline where estimation is optimized independently, followed by simultaneous prediction and planning. 
The factor graph structure (\figref{fig:system_factor_graph}) and covariance parameters remain identical across all configurations for fairness of comparison.

\begin{figure}[b]
    \centering
    \vspace{-2mm}
  \includegraphics[width=\linewidth]{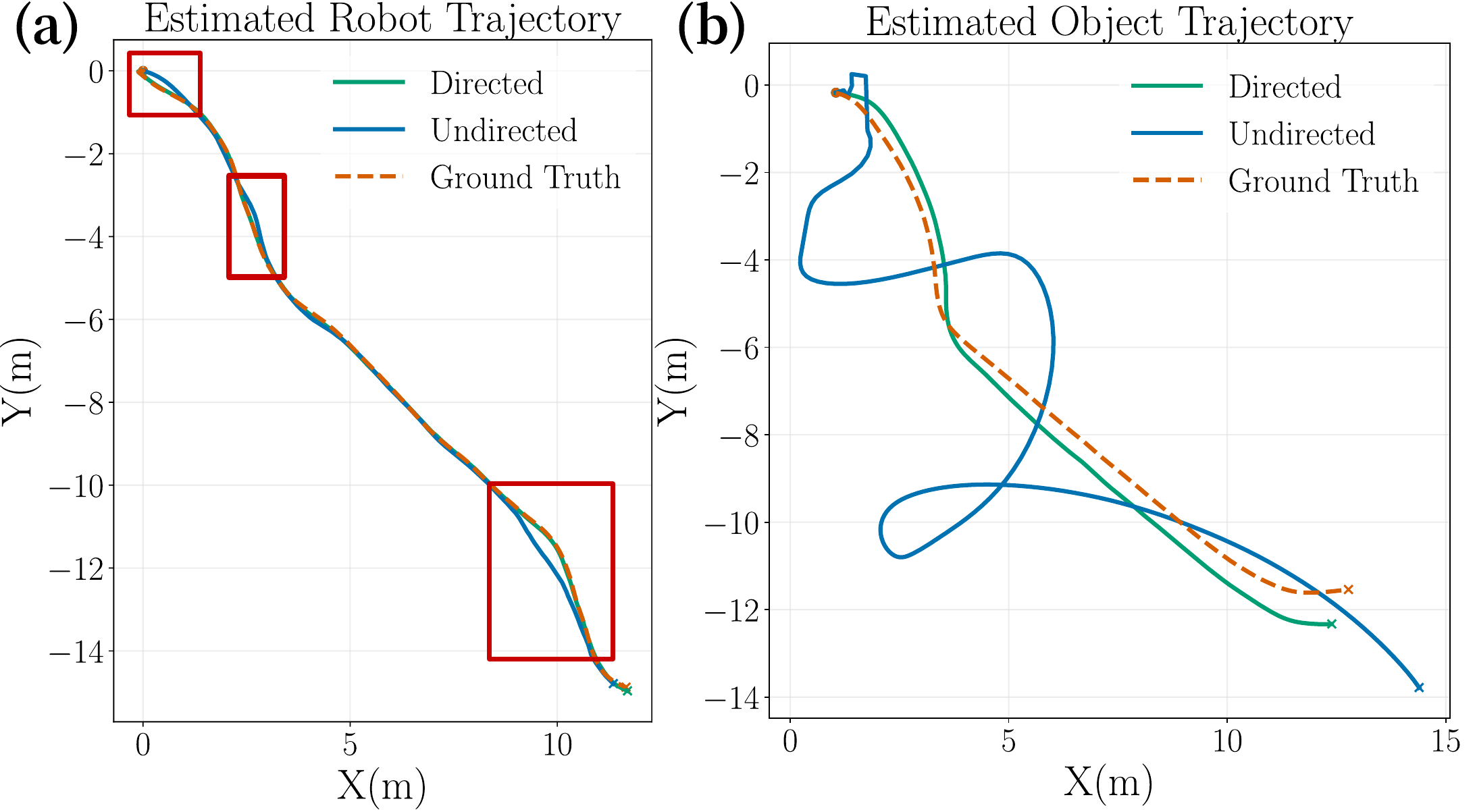}
   \caption{\small{Estimated trajectories at the end of navigation. Paths do not intersect as the object travels faster than the ego platform. In \textbf{(a)}, deviations from ground truth are highlighted with a red box, while in \textbf{(b)} the impact of undirected factors is evident.}}
    \label{fig:icra_2026_estimated_traj}
\end{figure}

\subsection{Simulation Setup}

Experiments are conducted in a simulated warehouse using Gazebo (\figref{fig:front_graphic}) with robots controlled via ROS2. The framework is optimized using a fixed-lag Levenberg–Marquardt implementation in GTSAM~\cite{gtsam}. At each time-step $k$, newly added factors and measurements update the optimizer, and the resulting control command $\acc{}{k}$ is executed.

The ego robot is a differential drive equipped with an Intel RealSense D435 RGB-D camera, with Gaussian noise ($\sigma=\SI{0.1}{\meter}$) added to per-pixel depth measurements. 
Objects are only detected when they appear in the camera field of view.
The static environment is mapped offline using SLAM Toolbox~\cite{macenski2021joss} and converted to a 2D ESDF. Low-frequency GPS-like pose measurements are provided to maintain map alignment.

A global reference path for the ego robot is generated via the Nav2 stack~\cite{macenski2023nav2} using Dijkstra’s algorithm, designed to pass near static obstacles to elicit local avoidance.
The local planner uses a horizon of $N=30$ with $\Delta t = \SI{0.1}{\second}$.
In all dynamic scenarios, moving obstacles are differential drive robots using the Nav2 MPPI controller. 
While these agents follow predefined trajectories in~\secref{sec:exp_DynoJEPP}, they are equipped with LiDAR in~\secref{sec:exp_CDynoJEPP} to detect the ego robot and perform \textit{reactive local avoidance}.

\subsection{DynoJEPP - Navigation}
\label{sec:exp_DynoJEPP}

\begin{table}
\footnotesize
\centering
\setlength{\tabcolsep}{1.5pt}
% \caption{\small{Quantitative results for $10$ navigation experiments in Gazebo. Experiments labeled \textbf{S} are in static environments and \textbf{D} in dynamic environments. At each time step ($\Delta t = 0.1$s), the \textbf{Motion Error (ME)}~\cite{morris2025dynosam} evaluates the estimated object trajectory available until that time step, and the table reports the mean over the entire run. \textbf{Goal} indicates whether the robot reached the goal successfully and \textbf{Steps} shows the number of time steps until the goal was reached or a collision occurred. All methods use the same factor graph structure (see~\figref{fig:system_factor_graph}). \textbf{Undirected} propagates information bidirectionally between modules, \textbf{Directed} uses directed factors, and \textbf{Decoupled} optimizes estimation alone before simultaneous prediction and planning. Best results are marked in bold and second best with underscore.}}
\caption{\small{Results for $10$ navigation experiments in Gazebo comparing proposed \textit{Directed} factor graph with \textit{Undirected} variant and \textit{Decoupled} baseline. Best results are bolded and second best underlined. Red color indicates number of steps before collision.}}
\label{tab:experiments_quantitative}
\begin{tabular}{cccccccccccc}
\toprule 

 & & \multicolumn{10}{c}{\textbf{Experiment ID (Static (S) / Dynamic (D))}} \\
\cmidrule(lr){3-12}
 & DynoJEPP & $\mathbf{1}$(S) & $\mathbf{2}$(S) & $\mathbf{3}$(D) & $\mathbf{4}$(D) & $\mathbf{5}$(D) & $\mathbf{6}$(D) & $\mathbf{7}$(D) & $\mathbf{8}$(D) & $\mathbf{9}$(D) & $\mathbf{10}$(D)\\
\midrule
\midrule
%   & & S & S & D & D & D & D & D & D & D & D &  \\
% \midrule
% \parbox[t]{6mm}{\multirow{2}{*}{GPS}} & Hz & & $0.5$ & $10$ & $10$ & $2$ & $10$ & $0.5$ & $0.2$ & $1$ & - & - &  \\
% & $\sigma$ & & $1e-1$ & $1e-2$ & $1e-2$ & $1e-1$ & $1e-2$ & $1e-1$ & $1e-2$ & $1e-2$ & - & - &  \\
% \midrule
\parbox[t]{2mm}{\multirow{3}{*}{\rotatebox[origin=c]{90}{$\text{ME}_r$(\si{\degree})}}} & Undirected & n/a & n/a & $5.12$ & $4.21$ & $2.51$ & $3.94$ & $0.79$ & $2.11$ & $1.19$ & $13.38$ \\
& Directed & n/a & n/a & $\underline{0.37}$ & $\mathbf{1.70}$ & $\mathbf{1.63}$ & $\underline{0.87}$ & $\underline{0.39}$ & $\underline{0.82}$ & $\mathbf{0.57}$ & $\mathbf{0.47}$ \\
& Decoupled & n/a & n/a & $\mathbf{0.32}$ & $\underline{1.72}$ & $\underline{1.64}$ & $\mathbf{0.68}$ & $\mathbf{0.37}$ & $\mathbf{0.38}$ & $\underline{0.60}$ & $\underline{2.48}$ \\
\midrule
\parbox[t]{2mm}{\multirow{3}{*}{\rotatebox[origin=c]{90}{$\text{ME}_t$(\si{\meter})}}} & Undirected & n/a & n/a & $0.05$ & $0.10$ & $0.04$ & $0.09$ & $0.03$ & $0.04$ & $0.03$ & $0.16$ \\
 & Directed & n/a & n/a & $\mathbf{0.01}$ & $\mathbf{0.02}$ & $\mathbf{0.01}$ & $\mathbf{0.01}$ & $\mathbf{0.02}$ & $\mathbf{0.01}$ & $\mathbf{0.01}$ & $\mathbf{0.02}$ \\
 & Decoupled & n/a & n/a & $\mathbf{0.01}$ & $\mathbf{0.02}$ & $\underline{0.02}$ & $\mathbf{0.01}$ & $\mathbf{0.02}$ & $\mathbf{0.01}$ & $\mathbf{0.01}$ & $\underline{0.05}$ \\
\midrule
\parbox[t]{2mm}{\multirow{3}{*}{\rotatebox[origin=c]{90}{Success}}} & Undirected & \xmark & \cmark & \xmark & \xmark & \xmark & \xmark & \xmark & \xmark & \xmark & \xmark \\
& Directed & \cmark & \cmark & \cmark & \cmark & \cmark & \cmark & \cmark & \cmark & \cmark & \cmark \\
& Decoupled & \cmark & \cmark & \cmark & \cmark & \cmark & \cmark & \cmark & \cmark & \cmark & \cmark \\
\midrule
\parbox[t]{2mm}{\multirow{3}{*}{\rotatebox[origin=c]{90}{Steps}}} & Undirected & \textcolor{red}{$180$} & $\mathbf{184}$ & \textcolor{red}{$32$} & \textcolor{red}{$36$} & \textcolor{red}{$128$} & \textcolor{red}{$30$} & \textcolor{red}{$51$} & \textcolor{red}{$56$} & \textcolor{red}{$71$} & \textcolor{red}{$84$} \\
& Directed & $\underline{233}$ & $\mathbf{184}$ & $\mathbf{92}$ & $\underline{235}$ & $\mathbf{239}$ & $\mathbf{106}$ & $\mathbf{167}$ & $\mathbf{111}$ & $\underline{134}$ & $\underline{201}$ \\
& Decoupled & $\mathbf{205}$ & $189$ & $\underline{94}$ & $\mathbf{234}$ & $\underline{244}$ & $\underline{109}$ & $\underline{185}$ & $\underline{121}$ & $\mathbf{125}$ & $\mathbf{156}$ \\
\bottomrule
\end{tabular}
\vspace{-4mm}
\end{table}

\begin{figure}[t] % Use figure* for spanning both columns
    \centering
     \includegraphics[width=0.9\linewidth]{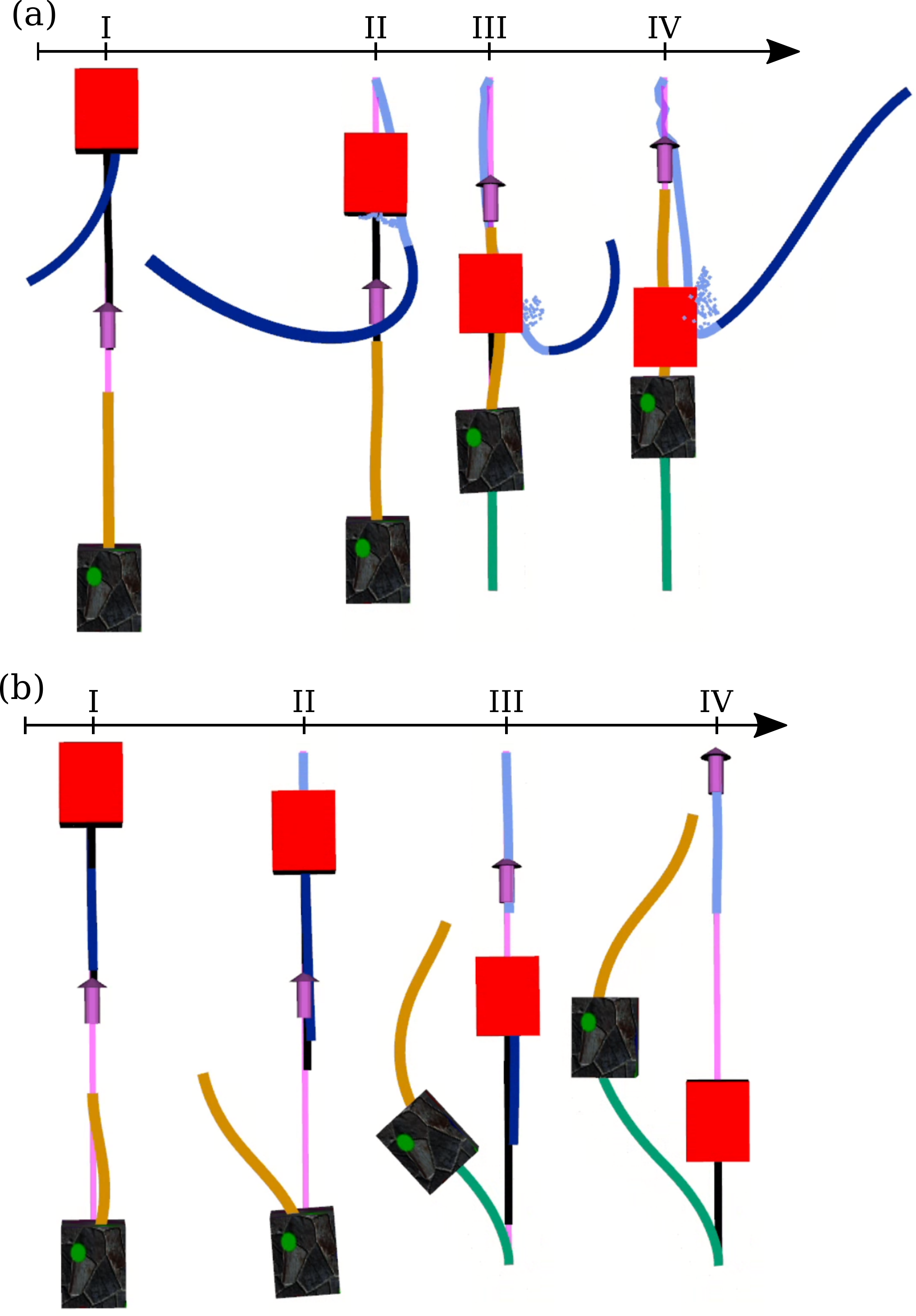}
    % \caption{Dynamic object avoidance, demonstrating collision as a consequence of overly optimistic planning with \textit{undirected} factors in \textbf{(a)} compared to \textit{directed} factors in \textbf{(b)} which generates safe plans around the moving object in accordance with the latest estimation. 
    % The ego robot is depicted as a black block and the dynamic object in red. Trajectories are colored as follows: ego estimation (green), object estimation (light blue), object prediction (dark blue), ego local plan (orange), and the global plan pink with the local goal as an arrow.}
    \caption{\small{Dynamic object avoidance comparing \textbf{(a)} \textit{Undirected} and \textbf{(b)} \textit{Directed} configurations.
    Undirected factors lead to collisions due to overly optimistic planning. 
    Colors: ego robot (\textcolor{EgoRobot}{black box}), ego estimation (\textcolor{EgoEstimation}{green}), ego local plan (\textcolor{EgoPlan}{orange}), global plan (\textcolor{GlobalPlan}{pink}), local goal (\textcolor{LocalGoal}{arrow}), dynamic object (\textcolor{DynamicObject}{red box}), object estimation (\textcolor{ObjectEstimation}{light blue}), object prediction (\textcolor{ObjectPrediction}{dark blue}), and ground-truth object local plan (black).}}
    \label{fig:dynamic_object_avoidance_undirected}
    \label{fig:dynamic_object_avoidance_directed}
    \label{fig:side_by_side}
    \label{fig:qualitative_dynojepp}
\vspace{-4mm}
\end{figure}

We evaluate the impact of directed factors across $10$ runs in static (\textbf{S}) and dynamic (\textbf{D}) environments, reporting navigation success (\textbf{Success}), time-steps to reach the goal or collide (\textbf{Steps}), and the Motion Error (\textbf{ME})~\cite{morris2025dynosam} of object trajectory estimations averaged over all time-steps, in \tabref{tab:experiments_quantitative}. 
We compare \textit{Undirected}, \textit{Directed}, and \textit{Decoupled} configurations using the same graph structure. Robot pose errors are omitted, as localization and measurement factors tightly constrain the ego robot pose.

In static environments, prediction factors are absent, yet planning and estimation still interact through the motion model factor linking robot poses $\mathbf{X}_{k}$ and $\mathbf{X}_{k+1}$. Proximity to static obstacles creates high costs in the planning component which, in the \textit{Undirected} case, propagates back to corrupt the estimated trajectory. In cluttered scenarios like Experiment $1$, this corruption leads to poor localization and infeasible plans, causing the \textit{Undirected} configuration to crash.

In dynamic environments, directed factors are essential to prevent planning costs from unrealistically influencing object predictions. 
As shown in \figref{fig:dynamic_object_avoidance_undirected}(a), corresponding to Experiment $3$, undirected predictions are unrealistic and adapt to the overly optimistic robot's plan. This leads the \textit{Undirected} method to generate unsafe trajectories, failing all attempted test cases.
\figref{fig:icra_2026_estimated_traj} shows that undirected factors degrade both robot and object trajectory estimates, with object motion most affected. Conversely, the \textit{Directed} method preserves accurate estimations and realistic predictions while ensuring planned trajectories safely avoid obstacles (\figref{fig:dynamic_object_avoidance_directed}(b)).
As shown in \tabref{tab:experiments_quantitative}, \textit{Directed} and \textit{Decoupled} achieve similar accuracy, significantly outperforming \textit{Undirected}. This demonstrates that our formulation maintains baseline estimation quality while enabling simultaneous optimization.

However, differences in initialization and the optimization process generate non-identical control commands, placing the robot in distinct situations. 
For example, in Experiment $10$, the \textit{Directed} method waits for the dynamic object to pass, whereas \textit{Decoupled} accelerates to move in front of it. 
While a detailed control analysis is beyond the scope of this work, these variations result in distinct but viable behaviors without one configuration consistently outperforming the other.
Overall, directed factors are critical for robust navigation in both static and dynamic scenarios.

\subsection{Cooperative DynoJEPP - Navigation}
\label{sec:exp_CDynoJEPP}

\begin{figure}[!h]
    \centering
    \includegraphics[width=0.95\columnwidth]{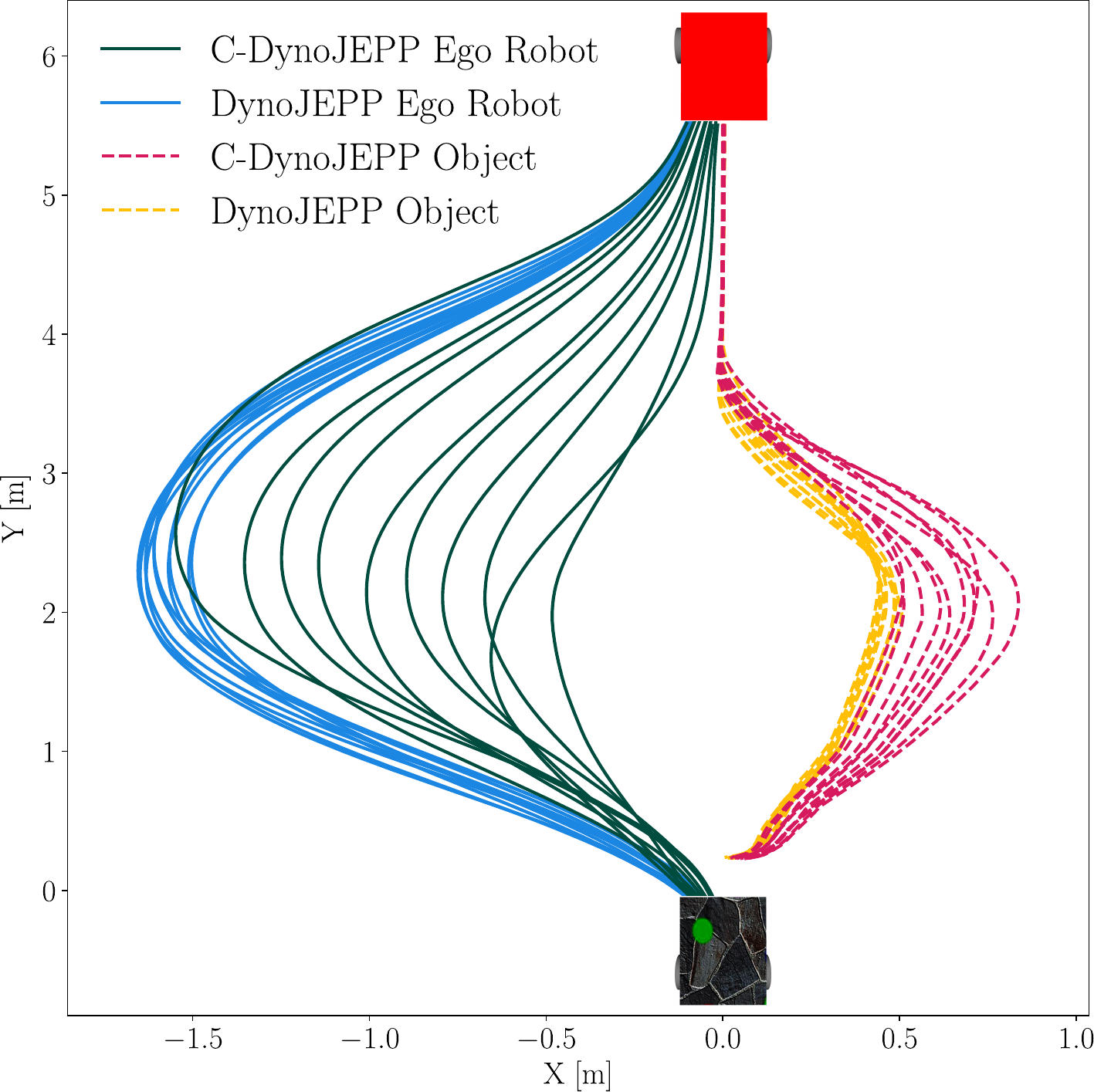}
    \caption{\small{Ground truth trajectories of the ego robot (black) and dynamic object (red) under DynoJEPP and C-DynoJEPP. Cooperation weights vary across $10$ runs shown for cooperative method.}}
    % \caption{\small{Ground truth trajectories of the ego robot and dynamic object under DynoJEPP and C-DynoJEPP (10 runs each). Cooperation weights vary in C-DynoJEPP.}}
    % 1e-4, 1e-3, 1e-2, 5e-2, 1.5e-1, 2e-1, 2.5e-1, 2.3e-1, 5e-1, 3.6e-1
% Num of steps to complete CDynoJEPP: 84, 82, 82, 82, 86, 87, 90, 89, 94, 89
% Num of steps to complete DynoJEPP: 92, 100, 97, 98, 97, 101, 94, 92, 98, 92
    \vspace{-4mm}
    \label{fig:qualitative_cdynojepp}
\end{figure}

We compare C-DynoJEPP to DynoJEPP using the same scenario as in Experiment $3$, but with the dynamic object exhibiting \textit{reactive local avoidance}.
\figref{fig:qualitative_cdynojepp} displays ground truth trajectories across $10$ runs for each method.
While DynoJEPP produces conservative paths, C-DynoJEPP enables more assertive navigation by incorporating cooperation weights into the prediction, thereby influencing the local plan. 
Quantitatively, the fastest C-DynoJEPP configuration reached the goal in $82$ steps, outperforming the fastest DynoJEPP run ($92$ steps) and its $96$-step average.
These results demonstrate the versatility of the cooperative framework and its utility in scenarios requiring interactive navigation.

% Under DynoJEPP the ego robot consistently follows conservative paths. 
% In C-DynoJEPP, each trajectory corresponds to a different constant cooperation level, resulting in variations in the predicted trajectories and the robot’s local plan. 
% The fastest C-DynoJEPP configuration reaches the goal in $72$ time steps, while DynoJEPP requires $95$ steps on average.
% These results demonstrate the versatility of C-DynoJEPP and its potential as a useful option in scenarios where more aggressive or cooperative navigation is desired.

\section{Conclusion and Future Work}

This paper introduces \textit{DynoJEPP}, a factor-graph-based framework for simultaneous estimation, $\SE$ motion prediction, and local planning in dynamic environments.
We proposed a novel \textit{directed factor} to enforce causal dependencies, effectively enabling control over the information flow among architecture components.
Experimental validation in dynamic scenarios confirms that controlling this information flow is critical for successful navigation, allowing the framework to safely avoid both static and dynamic obstacles.
Our analysis indicates that while estimation does not inherently benefit from joint optimization, prediction and planning can benefit from it, particularly when modeling interactive dynamic object behavior, as in the \textit{C-DynoJEPP} extension.
Future research will focus on developing an efficient, incremental, and multi-threaded implementation of DynoJEPP to achieve real-time performance. This involves leveraging incremental dynamic SLAM~\cite{Morris2025ral_dynosam}, online mapping~\cite{Wang2025icra_dynorecon}, and benchmarking against industry-standard navigation frameworks such as Nav2~\cite{macenski2023nav2}.

\balance
\bibliographystyle{IEEEtran}
\bibliography{./IEEEabrv, ./refs/bibliography}

\end{document}